\documentclass{article}

\PassOptionsToPackage{numbers, compress}{natbib}
 \usepackage[preprint]{neurips_2026}

\usepackage[utf8]{inputenc} 
\usepackage[T1]{fontenc}    
\usepackage{hyperref}       
\usepackage{url}            
\usepackage{booktabs}       
\usepackage{amsfonts}       
\usepackage{nicefrac}       
\usepackage{microtype}      
\usepackage{xcolor}         
\usepackage{graphicx}
\usepackage{pifont}
\usepackage{amssymb}
\usepackage{bbding}
\usepackage{multirow}
\usepackage{makecell}
\usepackage{threeparttable}
\usepackage{float}
\usepackage{subcaption}
\usepackage{booktabs}

\newcommand{\yx}[1]{#1}
\newcommand{\zyx}[1]{#1}

\definecolor{myred}{rgb}{1, 0.7, 0.7}
\newcommand{\reducedstrut}{\vrule width 0pt height 1.05\ht\strutbox depth 1.0\dp\strutbox\relax}
\newcommand{\sota}[1]{%
  \begingroup
  \setlength{\fboxsep}{0pt}%
  \colorbox{myred}{\reducedstrut#1\/}%
  \endgroup
}

\usepackage{tcolorbox}
\tcbuselibrary{breakable,listings}

\title{\yx{Toward Visually Realistic Simulation: A Benchmark for Evaluating Robot Manipulation in Simulation}}

\author{
\textbf{Yixin Zhu}$^{1}$ \quad
\textbf{Zixiong Wang}$^{2}$ \quad
\textbf{Jian Yang}$^{1}$ \vspace{0.1em} \\
\textbf{Jin Xie}$^{1}$ \quad
\textbf{Jingyi Yu}$^{3}$ \quad
\textbf{Jiayuan Gu}$^{3}$ \quad
\textbf{Beibei Wang}$^{1,*}$ 
\vspace{0.4em} \\
$^{1}$Nanjing University \quad
$^{2}$Nankai University \quad
$^{3}$ShanghaiTech University 
\vspace{0.4em} \\
$^{*}$ Corresponding Author
}

\begin{document}

\maketitle

\begin{figure}[htbp]
    \centering
    \includegraphics[width=1\linewidth]{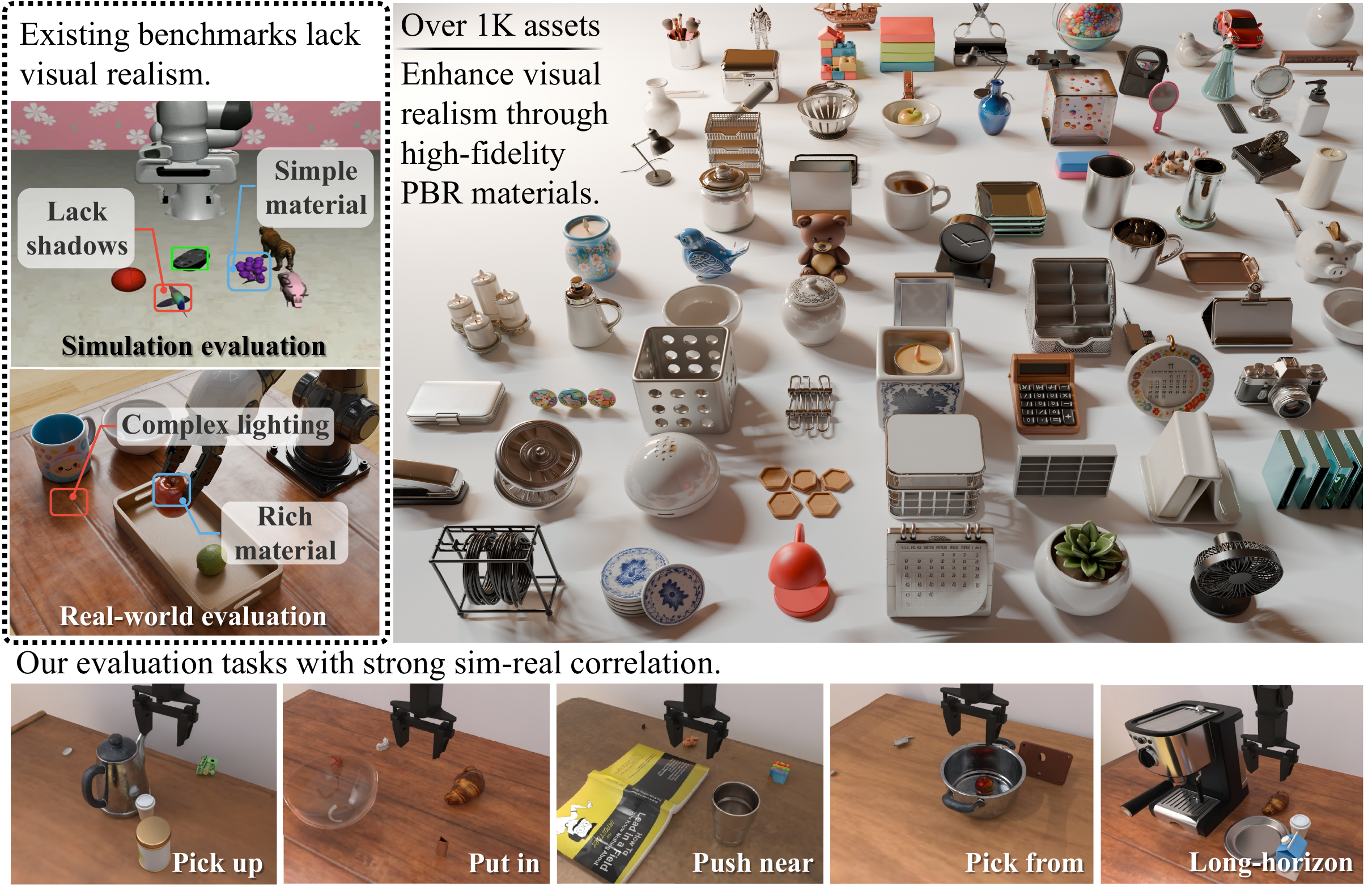}
    \caption{
    We introduce \textbf{VISER}, a \textbf{VIS}ually realistic benchmark for \textbf{E}valuating \textbf{R}obot manipulation in simulation. 
    VISER features a high-fidelity dataset of over 1,000 3D assets with PBR materials, along with 3D scenes constructed from these assets through curated layouts or generation.
    We show that enhancing visual realism greatly boosts the consistency between simulations and the real world.
    }
    \label{fig:teaser}
\end{figure}

\begin{abstract}
\yx{
Reliable simulation evaluation of robot manipulation policies serves as a high-fidelity proxy for real-world performance.
Although existing benchmarks cover a wide range of task categories, they lack visual realism, creating a large domain gap between simulation and reality. 
\zyx{This undermines the reliability of simulation-based evaluation in predicting real-world performance.}
To mitigate the sim-to-real visual gap, we conduct a systematic analysis to isolate the effects of lighting and material. Our results show that these factors play a critical role in geometric reasoning and spatial grounding, yet are largely overlooked in existing benchmarks. 
\zyx{Motivated by the analysis}, we propose VISER, a visually realistic benchmark for evaluating robot manipulation in simulation. \zyx{VISER features a high-fidelity dataset of over 1,000 3D assets with physically-based rendering (PBR) materials, along with 3D scenes created from these assets through curated layouts or generation.}
To this end, we propose an automated pipeline leveraging Multi-modal Large Language Models (MLLMs) for material-aware part segmentation and material retrieval, enabling scalable generation of physically plausible assets.
Building on the high-fidelity 3D asset dataset, we construct diverse evaluation tasks, such as grasping, placing, and long-horizon tasks, enabling scalable and reproducible assessment of Vision-Language-Action (VLA) models. 
\zyx{Our benchmark shows a strong correlation between simulation and real-world performance, achieving an average Pearson correlation coefficient of 0.92 across different policies.}
}


\end{abstract}

\section{Introduction}

Robot manipulation policies \citep{kim2024openvlaopensourcevisionlanguageactionmodel,octomodelteam2024octoopensourcegeneralistrobot,black2026pi0visionlanguageactionflowmodel} have recently achieved strong generalization across tasks and environments. As these policies become increasingly capable, reliably evaluating their generalization and robustness has become critically important, yet remains a major bottleneck. Although real-world evaluation is considered more reliable, its high cost and scalability limitations necessitate the use of simulation-based evaluation as a promising alternative. A key challenge lies in the sim-to-real gap, which undermines the reliability of simulation as a proxy for real-world performance. \zyx{The domain gap between simulated and real-world environments can be divided into the visual and physical discrepancies. While the latter has been effectively mitigated by system identification, the former remains underexplored in the context of robot manipulation policy.}

\zyx{
Existing benchmarks~\citep{liu2023liberobenchmarkingknowledgetransfer,zhang2024vlabenchlargescalebenchmarklanguageconditioned,mees2022calvinbenchmarklanguageconditionedpolicy,szot2022habitat20traininghome,li2024evaluatingrealworldrobotmanipulation} mainly focus on domain randomization to mitigate the visual gap. However, they largely overlook an important but underexplored visual realism, oversimplifying the modeling of lighting and material properties.
}
\yx{Specifically, although SimplerEnv~\citep{li2024evaluatingrealworldrobotmanipulation} bridges the visual gap, the reliance on ``green-screening'' inevitably introduces physically implausible lighting artifacts (e.g., missing shadows and incorrect reflections).} As a result, existing benchmarks may yield a mismatch evaluation between simulation and the real world.

\yx{While visual realism is key to mitigating this gap, the specific visual factors that matter most in simulation-based evaluation remain underexplored. Therefore, we conduct a systematic study of the sim-to-real visual gap, isolating the effects of light and material. Our results show that the specular highlights in material and shadows from light significantly impact both geometric reasoning and spatial grounding in Vision-Language-Action (VLA) models, yet are largely neglected in existing benchmarks.}
This motivates us to introduce \textbf{VISER}, a \textbf{VIS}ually realistic benchmark for \textbf{E}valuating \textbf{R}obot manipulation in simulation. It comprises a photorealistic 3D asset dataset for robot manipulation evaluation, and a suite of standardized evaluation tasks. Our dataset contains over 1K high-fidelity 3D assets with clean PBR materials.
To construct this 3D asset dataset, we propose an automated pipeline leveraging Multi-modal Large Language Models (MLLMs) for material-aware part segmentation and retrieval, ensuring physically plausible appearance. While existing datasets~\citep{chen2025robotwin20scalabledata,wang2026manitwinscalingdatagenerationreadydigital} suffer from baked-lighting artifacts due to 3D generation approaches. 
\zyx{Built upon this 3D asset dataset, we establish a standardized simulation evaluation with diverse tasks, including 14 curated tasks, 8 reconstructed tasks from real-world scenes, and a set of generated tasks. We demonstrate a strong correlation between policy performance in our simulation and the real world, achieving an average Pearson correlation coefficient of 0.92 with real-world performance across different policies, significantly outperforming existing baselines.}
\yx{
To summarize, our contributions are as follows:
\begin{itemize}
    \item We identify visual realism (specular highlights in material and shadows from light) as a critical but underexplored factor in simulation-based robot evaluation, supported by systematic analysis.
    \item We construct a large-scale dataset with over 1K high-fidelity 3D assets featuring PBR materials without baked-lighting artifacts, enabling realistic simulation-based evaluation.
    \item We propose a benchmark for robot manipulation with high visual realism, demonstrating strong sim-to-real correlation.
\end{itemize}
}

\section{Related Work}

\paragraph{Vision-Language-Action models.}
With advances in Vision-Language models (VLMs)~\citep{Qwen2.5-VL,singh2025openaigpt5card,geminiteam2025geminifamilyhighlycapable}, VLAs~\citep{octomodelteam2024octoopensourcegeneralistrobot, kim2024openvlaopensourcevisionlanguageactionmodel, black2026pi0visionlanguageactionflowmodel,geminiroboticsteam2025geminirobotics15pushing,intelligence2025pi05visionlanguageactionmodelopenworld} have recently attracted significant attention due to their scalability. \zyx{Current VLAs trained on large-scale datasets, such as OpenVLA~\citep{kim2024openvlaopensourcevisionlanguageactionmodel}, $\pi_0$~\citep{black2026pi0visionlanguageactionflowmodel}, and $\pi_{0.5}$~\citep{intelligence2025pi05visionlanguageactionmodelopenworld}, present strong generalization from within-domain tasks to out-of-distribution scenarios.} However, the rapid advances and the need for extensive testing make the real-world evaluation challenging. \zyx{In contrast, simulated evaluation is a promising alternative, offering not only efficiency but also enabling safe assessment of performance in novel environments or with unfamiliar objects.} To achieve an efficient and reliable simulated evaluation, we construct an evaluation framework based on high-quality 3D assets and photorealistic rendering.
\vspace{-3mm}
\paragraph{\yx{Benchmarks and datasets for robot manipulation.}}
Real-world datasets~\citep{walke2024bridgedatav2datasetrobot,fu2024mobilealohalearningbimanual,open_x_embodiment_rt_x_2023} provide a diverse range of high-quality demonstrations that are essential for training robust robotic manipulation policies, but are difficult to replicate for evaluation. Simulated evaluation is an efficient alternative to real-world evaluation. A wide range of simulation benchmarks~\citep{li2024evaluatingrealworldrobotmanipulation,liu2023liberobenchmarkingknowledgetransfer,zhang2024vlabenchlargescalebenchmarklanguageconditioned,mees2022calvinbenchmarklanguageconditionedpolicy,zheng2024robocasbenchmarkroboticmanipulation} have been proposed to facilitate evaluation. Most simulated benchmarks are built on MuJoCo~\citep{todorov2012mujoco}, which provides plausible physical simulation and fast rasterization, but they lack visual fidelity. \zyx{Similarly, current robotics datasets, such as RoboTwin~\citep{mu2025robotwindualarmrobotbenchmark,chen2025robotwin20scalabledata} and ManiTwin~\citep{wang2026manitwinscalingdatagenerationreadydigital}}, usually utilize assets generated by 3D generation models, which, despite their scalability, lack the high-fidelity textures necessary for real-world representation. This visual discrepancy makes these evaluation frameworks unreliable; for instance, the policy may fail in the real-world environment with complex materials and illumination, even when it behaves perfectly in simulation environments.

\begin{table*}[htbp]
\vspace{-3mm}
\centering
\begin{threeparttable}
\footnotesize
\setlength{\tabcolsep}{4pt}
\renewcommand{\arraystretch}{1.1}
\caption{
Comparison of simulation-based evaluation benchmarks or datasets for robot manipulation.
\textbf{Soft shadow}: Natural soft shadows in rendering.
\textbf{Specular}: Proper specular highlights in rendering.
\textbf{Clean PBR}: materials without baked lighting artifacts.
\textbf{Sim-to-Real Corre.}: validated correlation with real-world performance.
\textbf{Asset Cat.}: The categories of assets.
\textbf{\#Asset}: The number of assets.
\textbf{\#Task}: Ther number of tasks.
While prior benchmarks emphasize task diversity or domain randomization, they largely overlook material-level visual realism.
}
\begin{tabular}{lcccccccc}

\toprule
\textbf{B/D} 
& \textbf{Soft shadow} 
& \textbf{Specular} 
& \textbf{Clean PBR} 
& \textbf{Sim-to-Real Corr.} 
& \textbf{Asset Cat.} 
& \textbf{\#Asset}
& \textbf{\#Task} \\
\midrule
LIBERO~\citep{liu2023liberobenchmarkingknowledgetransfer}
& \textcolor{red}{\XSolidBrush} & \textcolor{red}{\XSolidBrush} & \textcolor{red}{\XSolidBrush} & \textcolor{red}{\XSolidBrush} & 51 & 75 & 130\\

CALVIN~\citep{mees2022calvinbenchmarklanguageconditionedpolicy}
& \textcolor{red}{\XSolidBrush} & \textcolor{red}{\XSolidBrush} & \textcolor{red}{\XSolidBrush} & \textcolor{red}{\XSolidBrush} & 5 & 30 & 34\\

ManiSkill~\citep{mu2021maniskillgeneralizablemanipulationskill}
& \textcolor{teal}{\Checkmark} & \textcolor{teal}{\Checkmark} & \textcolor{red}{\XSolidBrush} & \textcolor{red}{\XSolidBrush} & 100 & 2,600 & 20\\

Behavior-1K~\citep{li2024behavior1khumancenteredembodiedai}
& \textcolor{teal}{\Checkmark} & \textcolor{teal}{\Checkmark} & \textcolor{red}{\XSolidBrush} &  \textcolor{orange}{\Checkmark} & -- & 10K & 1,000\\

Habitat 2.0~\citep{szot2022habitat20traininghome}
& \textcolor{red}{\XSolidBrush} & \textcolor{orange}{\Checkmark} & \textcolor{red}{\XSolidBrush} & \textcolor{red}{\XSolidBrush} & 46 & 169 & 3\\

SimplerEnv~\citep{li2024evaluatingrealworldrobotmanipulation}
& \textcolor{red}{\XSolidBrush} & \textcolor{orange}{\Checkmark} & \textcolor{red}{\XSolidBrush} & \textcolor{teal}{\Checkmark} & -- & 17 & 8\\

RoboCASA~\citep{robocasa2024} 
& \textcolor{teal}{\Checkmark} & \textcolor{teal}{\Checkmark} & \textcolor{red}{\XSolidBrush} & \textcolor{red}{\XSolidBrush} & 153 & 2,509 & 100\\

VLABench~\citep{zhang2024vlabenchlargescalebenchmarklanguageconditioned}
& \textcolor{red}{\XSolidBrush} & \textcolor{red}{\XSolidBrush} & \textcolor{red}{\XSolidBrush} & \textcolor{red}{\XSolidBrush} & 163 & 2,164 & 100\\

RoboTwin~\citep{mu2025robotwindualarmrobotbenchmark}
& \textcolor{orange}{\Checkmark} & \textcolor{teal}{\Checkmark} & \textcolor{red}{\XSolidBrush} & \textcolor{red}{\XSolidBrush} & 147 & 731 & 50\\

ManiTwin~\citep{wang2026manitwinscalingdatagenerationreadydigital}
& -- & -- & \textcolor{red}{\XSolidBrush} & \textcolor{red}{\XSolidBrush} & -- & 100K  & -- \\

\midrule
\textbf{Ours} 
&  \textcolor{teal}{\Checkmark} & \textcolor{teal}{\Checkmark} &  \textcolor{teal}{\Checkmark} &  \textcolor{teal}{\Checkmark} & 319 & 1,049 & 22+generated \\
\bottomrule
\end{tabular}

\begin{tablenotes}
\item \textcolor{teal}{\Checkmark} Yes/Supported \quad \textcolor{orange}{\Checkmark} Partial/Limited \quad  \textcolor{red}{\XSolidBrush} No/Not supported
\end{tablenotes}

\end{threeparttable}

\label{tab:benchmark_comparison}
\end{table*}

\vspace{-3mm}

\paragraph{3D asset generation.}
Mainstream 3D asset generation is divided into two separate processes: geometry and material generation. With the advances of diffusion models~\citep{lipman2023flowmatchinggenerativemodeling,peebles2023scalablediffusionmodelstransformers,esser2024scalingrectifiedflowtransformers}, generation methods based on score distillation~\citep{poole2022dreamfusiontextto3dusing2d} have been proposed to enable text-3D generation leveraging the prior in pre-trained 2D diffusion models. Driven by the growing availability of 3D datasets~\citep{chang2015shapenetinformationrich3dmodel,deitke2022objaverseuniverseannotated3d,liu2025uncommonobjects3d}, feedforward methods trained on native 3D data, such as CLAY~\citep{zhang2024claycontrollablelargescalegenerative}, Tripo~\citep{tochilkin2024triposrfast3dobject}, and LATTICE~\citep{lai2025latticedemocratizehighfidelity3d}, have achieved high-fidelity geometry generation. However, current material generation methods~\citep{he2025materialmvpilluminationinvariantmaterialgeneration,lai2025hunyuan3d25highfidelity3d,lai2025natexseamlesstexturegeneration,Hadadan_2025} still struggle to generate fully clean materials without lighting baking. On the other hand, material retrieval methods~\citep{zhang2025mapatextdrivenphotorealisticmaterial,fang2024makeitrealunleashinglargemultimodal,wang2026partuvpartbaseduvunwrapping} are proposed to enable plausible and clean material generation. 
These methods usually rely on segmentation approaches~\citep{xie2021segformersimpleefficientdesign,liu2025partfieldlearning3dfeature} and cluster masks of different parts heuristically. We achieve material-aware part segmentation by leveraging MLLM, enabling plausible material retrieval.

\section{Sim-to-Real Visual Gap Analysis} 

The domain gap between simulated and real-world environments can be divided into the visual and physical discrepancies. While the latter has been effectively mitigated by system identification, the former remains underexplored in the context of VLA. Given that perfectly replicating the complex material and lighting in the simulation is unfeasible, this section aims to isolate and identify the specific factors that significantly influence VLA performance. We first introduce our experimental setup in Sec.~\ref{sec:experiment_setup}, followed by a systematic analysis of the visual gap across two dimensions: specular highlights in material and shadows from light in Sec.~\ref {sec:analysis}. These results highlight the critical necessity of PBR materials and plausible light in bridging the sim-to-real gap for VLAs.

\subsection{Experimental Setup} \label{sec:experiment_setup}
Our experiments are conducted within the SimplerEnv framework~\citep {li2024evaluatingrealworldrobotmanipulation}, which is built upon the SAPIEN simulator~\citep{Xiang_2020_SAPIEN}. 
We replace original Simpler's ``green-screening'' with real assets, and use ray tracing to achieve photorealistic rendering.
The simulator achieves an average of 20 FPS on a single NVIDIA 4090 GPU, a frame rate we find sufficient for our high-fidelity benchmarking. We employ the WidowX robot and evaluate various policies using the BridgeData-v2 dataset~\citep{walke2024bridgedatav2datasetrobot}. Simulation reliability is assessed by measuring the alignment of success rates between the simulated and real-world environments. 

\begin{figure}[htbp]
    \centering
    \includegraphics[width=1\linewidth]{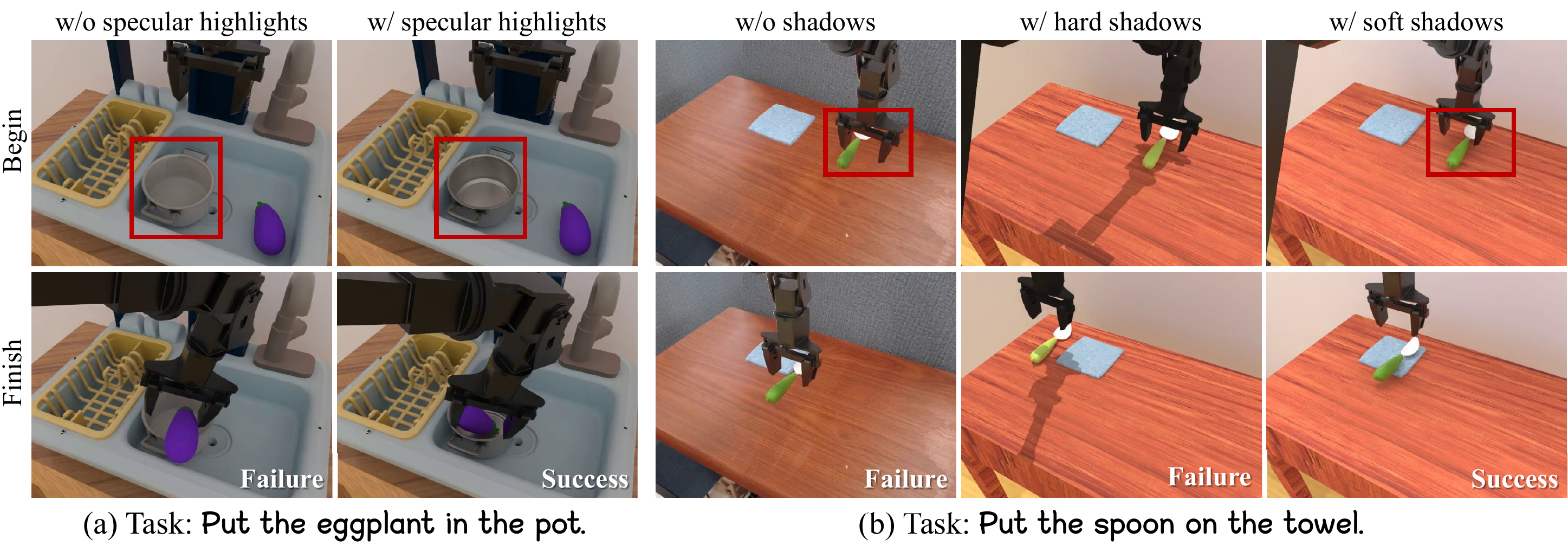}
    \vspace{-5mm}
    \caption{Visual factor analysis. (a) Specular improves geometry understanding. Without specular, the policy fails to understand the precise geometry of the pot, leading to failure in putting the eggplant in it. (b) Contact shadows improve spatial understanding. The policy fails to judge whether the spoon contacts the towel or not without shadows. Specifically, hard shadows also affect the performance of the VLAs. \zyx{More comparisons are available in Sec.~\ref{sec:correlation}.}}
    \label{fig:visual_analysis}
\end{figure}

\subsection{Visual Gap Analysis} \label{sec:analysis}

\paragraph{Specular highlights.} Specular highlights provide essential geometric priors that assist the VLA in accurately estimating the shape of target objects. To explore the factor of specular highlights, we set roughness to 1.0 and metallic to 0.0 to simulate a purely diffuse material without specular. As demonstrated in Fig.~\ref{fig:visual_analysis}(a), in the absence of these specular highlights, the VLA fails to localize the pot’s cavity, leading to task failure. In contrast, by rendering the pot with distinct specular, the model achieves a more robust 3D shape representation, thereby ensuring the successful execution of the pick-and-place task. As shown in Tab.~\ref{tab:highlight_analysis}, the success rate of the setup with specular is more consistent with the real world. Notably, the eggplant is always picked up successfully in both setups (100\%), and the difference mainly lies in the ``put into the pot'' step (drop from 90\% to 10\%), where complex geometry understanding is required.

\begin{table}[htbp]
    \centering
    \begin{minipage}[t]{0.56\textwidth}
        \centering
        \caption{Quantitative analysis of the sim-to-real gap caused by specular highlights.}
        \begin{tabular}{lcc}
            \toprule
            Setup & Grasp eggplant & Put eggplant into pot \\
            \midrule
            w/o specular & 100\% & 10\% \\
            w/ specular  & 100\% & 90\% \\
            Real world   & 100\% & 100\% \\
            \bottomrule
        \end{tabular}
        \label{tab:highlight_analysis}
    \end{minipage}
    \hfill 
    \begin{minipage}[t]{0.4\textwidth}
        \centering
        \caption{Quantitative analysis of the sim-to-real gap caused by cast shadows.}
        \begin{tabular}{lc} 
            \toprule
            Setup & Put spoon on towel \\
            \midrule
            w/o shadows     & 12\% \\
            w/ hard shadows & 0\% \\
            w/ soft shadows & 49\% \\
            Real world      & 42\% \\
            \bottomrule
        \end{tabular}
        \label{tab:shadow_analysis}
    \end{minipage}
\end{table}





\paragraph{Shadows.} Contact shadows provide essential depth and spatial cues, allowing the VLA to accurately perceive the relative distance between objects and surfaces. As illustrated in Fig.~\ref{fig:visual_analysis}(b), in the absence of shadows (applying ``green-screening''~\citep{li2024evaluatingrealworldrobotmanipulation}), the spoon appears to ``float'' above the table, obscuring its precise contact point. Consequently, the VLA fails to localize the spoon relative to the towel, resulting in an imprecise placement. This demonstrates that contact shadows are vital for grounding spatial relationships in visual reasoning, further underscoring the importance of PBR in generating physically plausible simulation environments. Quantitative results over 100 trials further corroborate this observation in Tab.~\ref{tab:shadow_analysis}: the success rate drops significantly. Moreover, the presence of hard shadows, whether caused by rasterization or strong directional light, leads to a significant degradation in VLA performance, with success rates dropping to nearly 0\%. Given that these sharp shadows are uncommon in real-world indoor settings, we conclude that they act as noise in synthetic environments, distorting the visual input and hindering VLA generalization.

\section{Method}
\begin{figure}[htbp]
    \centering
    \includegraphics[width=1\linewidth]{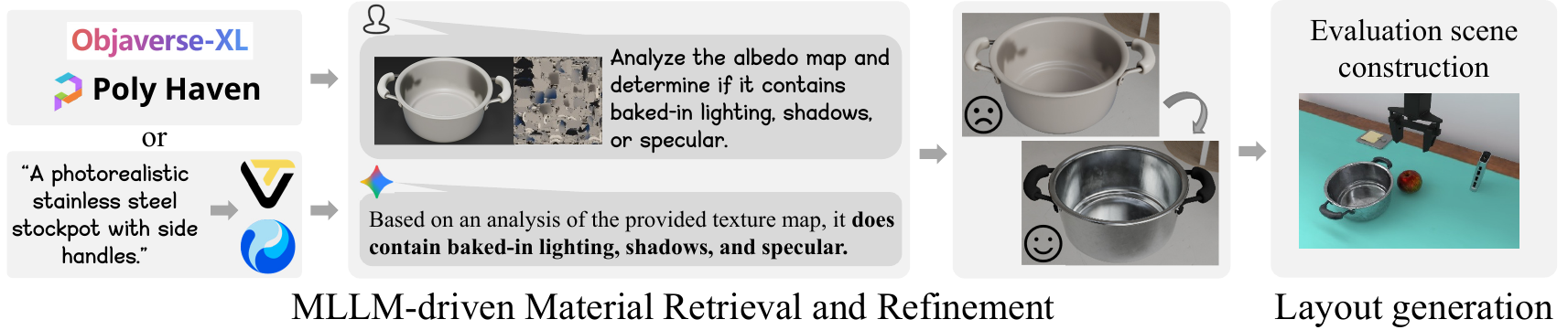}
    \caption{Pipeline of VISER construction. To enhance low-quality material, we propose an automated pipeline that uses an MLLM for material retrieval and refinement, resulting in physically plausible PBR materials. These high-quality assets are then used to construct diverse evaluation scenes via layout generation.}
    \label{fig:pipeline}
\end{figure}

\zyx{The above visual gap analysis motivates us to propose VISER, a visually realistic benchmark for evaluating robot manipulation in simulation. In this section, we introduce the pipeline to construct VISER, as shown in Fig.~\ref{fig:pipeline}. We first obtain textured meshes from public datasets or 3D generation approaches, and the material is refined through MLLM-driven material retrieval (Sec.~\ref{sec:retrieval}). Built upon the 3D asset dataset, evaluation scenes are constructed automatically by layout generation (Sec.~\ref{sec:layout}).}

\subsection{MLLM-driven Material Retrieval and Refinement} \label{sec:retrieval}
\begin{figure}
    \centering
    \includegraphics[width=1.0\linewidth]{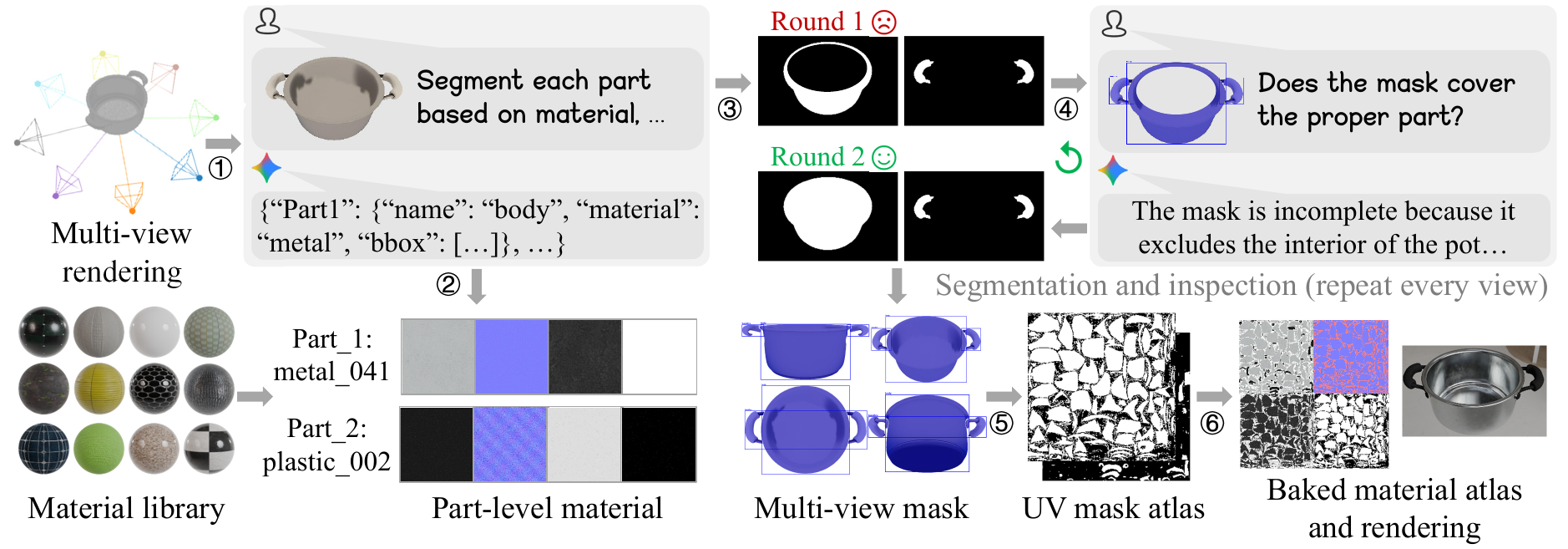}
    \caption{MLLM-driven material retrieval and refinement pipeline. \ding{172} Multi-view rendering and segmentation based on material types. \ding{173} Material retrieval according to the MLLM's segmentation. \ding{174} Part segmentation using SAM3 along with prompt and bounding boxes. \ding{175} Mask quality inspection and re-segmentation. \ding{176} Project mask from image space to UV space. \ding{177} Bake material into UV atlas.}
    \label{fig:retrieval_pipeline}
\end{figure}

We propose MLLM-driven material retrieval and refinement, as illustrated in Fig.~\ref{fig:retrieval_pipeline}. 
Unlike existing approaches that rely on heuristic-based mask refinement~\cite{fang2024makeitrealunleashinglargemultimodal}, our approach employs iterative MLLM-based inspection to ensure fine-grained part segmentation and precise material alignment.

To address viewpoint occlusions, we render objects from 32 perspectives and utilize an MLLM to perform material-aware segmentation. To maintain cross-view consistency, the MLLM first conducts a global scan to establish a unified part list, preventing granularity discrepancies across views. Each segmented part is then captioned by the MLLM to retrieve semantically matching materials from the material library~\cite{Vecchio_2024}. To enhance retrieval accuracy, we synthesize detailed descriptions for each library material using MLLM-rendered reference material balls.

For precise segmentation, we provide textual descriptions and MLLM-generated bounding boxes as prompts to SAM3. Recognizing that obscure views may degrade mask quality, we introduce a secondary MLLM-based evaluator to iteratively refine prompts. For instance, if a mask is incomplete (e.g., Round 1 in Fig.~\ref{fig:retrieval_pipeline}), the evaluator provides additional positive point prompts to SAM to ensure full coverage. Finally, these high-quality multi-view masks are projected into UV space, where they are integrated with the retrieved materials to generate PBR textures for individual components.

\subsection{Layout Generation} \label{sec:layout}
\begin{figure}
    \centering
    \includegraphics[width=1\linewidth]{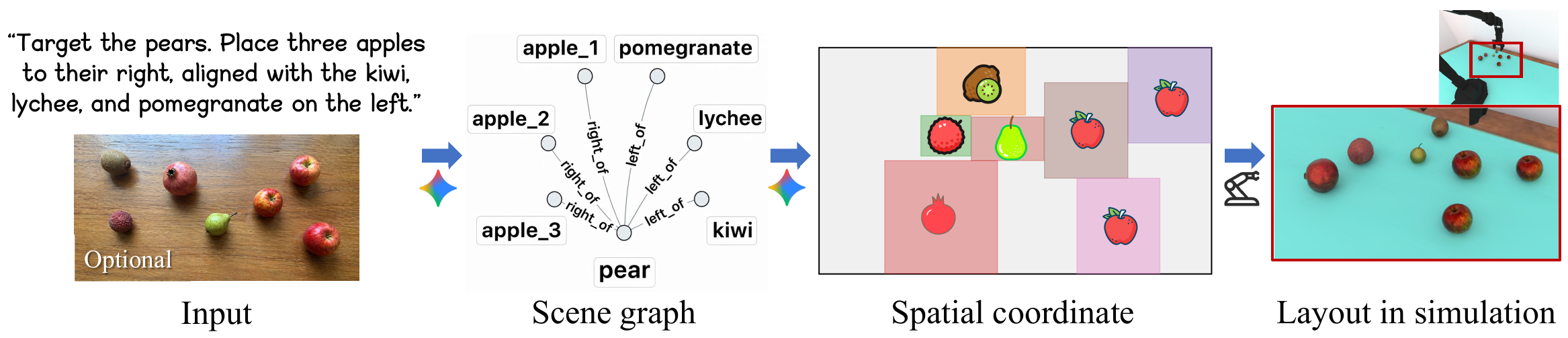}
    \caption{Layout generation. The LLM first extracts objects from the input to construct a scene graph representing the desired spatial arrangement. By integrating these object bounding boxes with the table's dimensions, the LLM calculates precise spatial coordinates and executes object placement within the simulation engine.}
    \label{fig:layout}
\end{figure}

Previous benchmarks, such as SimplerEnv~\citep{li2024evaluatingrealworldrobotmanipulation}, rely on manual design, which is both time-consuming and difficult to scale. To address this, we propose an automated approach to generate plausible layouts in the simulator based on user descriptions, as illustrated in Fig.~\ref{fig:layout}. Since simulated evaluation prioritizes scene diversity over pixel-perfect adherence to the input image, we utilize scene graphs to represent relative spatial relations, thereby avoiding the high computational overhead of optimization-based differentiable rendering. Given an input description or image, an LLM first extracts the objects and constructs a corresponding scene graph. Upon retrieving the assets, we provide the LLM with the objects’ bounding boxes and the table's dimensions to estimate precise spatial coordinates. By leveraging the table's fixed height, we reduce the problem to a 2D plane projection, which simplifies the task for the LLM compared to full 3D spatial reasoning. Finally, after validating the coordinates, the objects are instantiated in the simulation engine.

\section{VISER: A \underline{Vis}ually Realistic Benchmark for \underline{E}valuating \underline{R}obot Manipulation in Simulation}

\subsection{3D Asset Dataset}
\begin{figure}[htbp]
    \centering
    \includegraphics[width=1\linewidth]{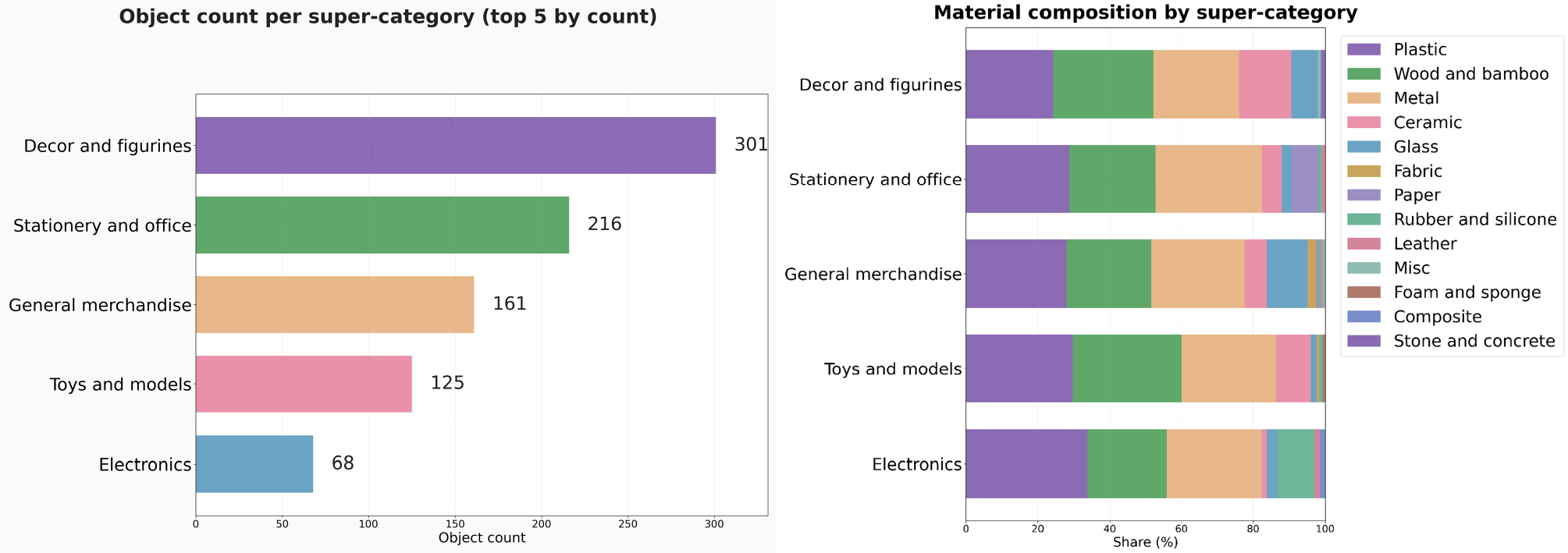}
    \caption{\yx{Overview of our asset dataset. Our asset dataset contains 12 super categories, 319 categories, and 1049 objects. Each super-category contains various material types, ensuring material diversity. We present the top five super-categories, and the full information is available in the supplementary.}}
    \label{fig:overview_dataset}
\end{figure}

\begin{figure}[htbp]
    \begin{minipage}[t]{0.48\linewidth}
        \centering
        \includegraphics[width=\linewidth]{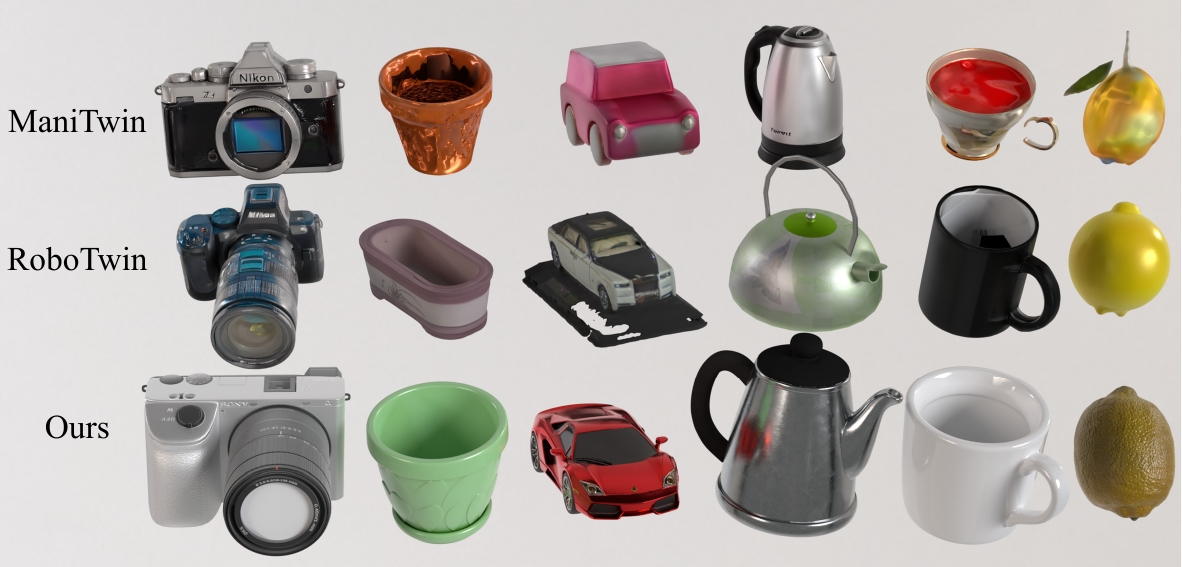}
        \caption{Comparison among asset datasets for robot simulation. Our asset dataset focuses on material quality and applies material retrieval to leverage high-fidelity material libraries.}
        \label{fig:dataset_comparison}
    \end{minipage}
    \hfill
    \begin{minipage}[t]{0.48\linewidth}
        \centering
        \includegraphics[width=\linewidth]{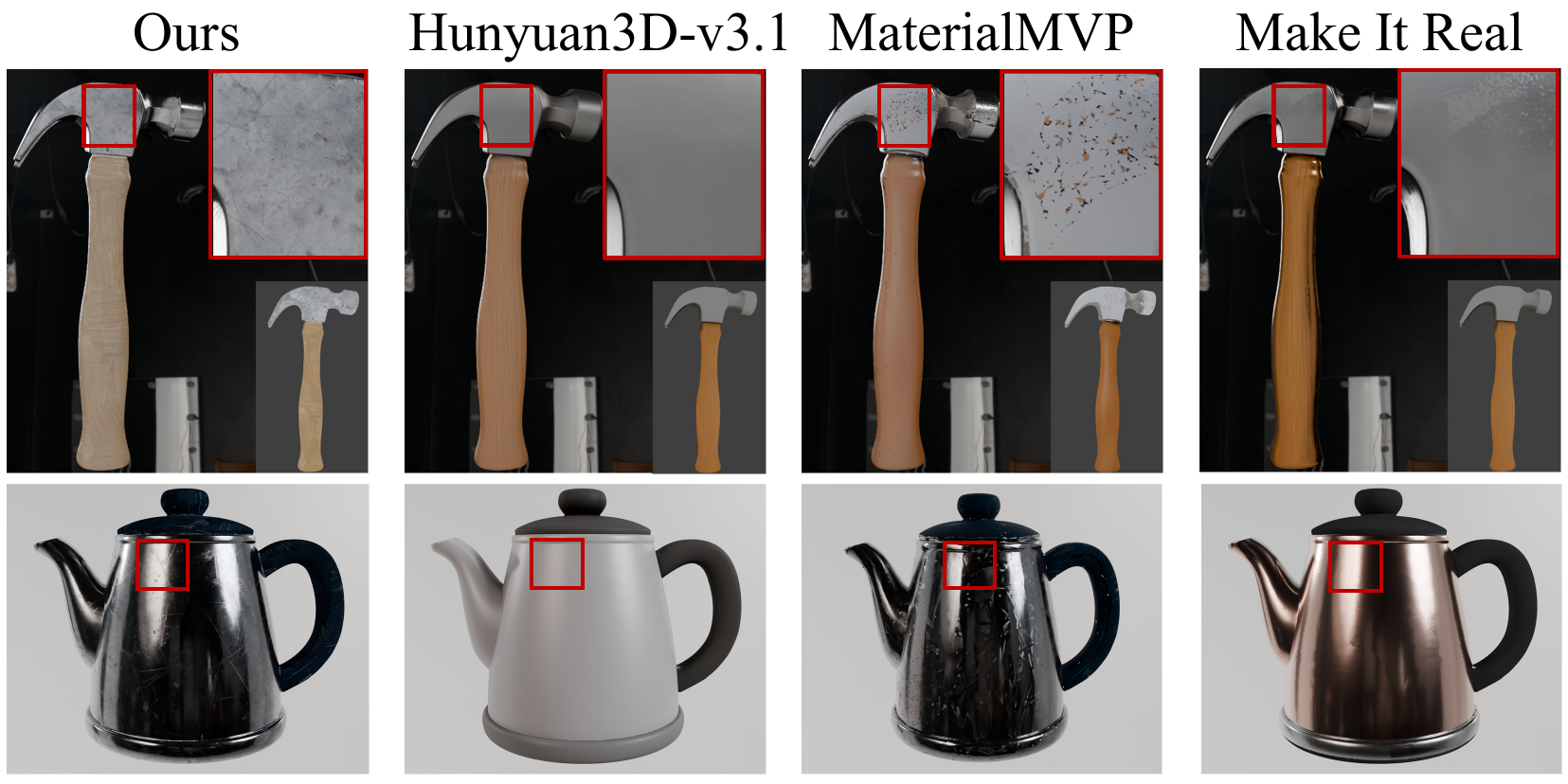}
        \caption{Qualitative comparison with state-of-the-art material generation and retrieval methods. Our approach excels in visual fidelity, capturing intricate details like metal scratches and stains.}
        \label{fig:mat_comparison}
    \end{minipage}
\end{figure}

Based upon MLLM-driven material retrieval and refinement, we construct a large-scale 3D asset dataset with high-fidelity PBR materials. An overview of our asset dataset is shown in Fig.~\ref{fig:overview_dataset}. Our dataset contains 12 super categories, 319 categories, and 1049 objects, covering the common table top merchandise, including stationery, electronics, toys, and so on. We present a qualitative comparison with the mainstream datasets in Fig.~\ref{fig:dataset_comparison}. Compared to RoboTwin~\citep{chen2025robotwin20scalabledata} and ManiTwin~\citep{wang2026manitwinscalingdatagenerationreadydigital}, our assets exhibit higher material fidelity, featuring rich surface details such as metallic scratches and ceramic-like textures, while fully avoiding light-baking artifacts.

\subsection{Benchmark Construction}
To ensure a comprehensive and reliable simulated evaluation, we design a diverse set of tasks, categorized into basic and long-horizon tasks. The basic category comprises five primitive skills: pick up, put in, push near, pick from, and open. Beyond evaluating mere action prediction, these tasks implicitly assess the model's spatial reasoning and embedded world knowledge. Unlike current benchmarks that rely on manually designed tasks~\citep{li2024evaluatingrealworldrobotmanipulation, zhang2024vlabenchlargescalebenchmarklanguageconditioned}, we procedurally sample contextually plausible objects and arrange them in realistic layouts tailored to each task type (detailed in Sec.~\ref{sec:layout}). Furthermore, to evaluate long-horizon reasoning capabilities, we introduce complex tasks driven by highly abstract goals (e.g., ``prepare the ingredients for an apple pie and clear the table'') rather than feeding explicit, step-by-step instructions to the VLA model. 

\yx{
Our benchmark contains three components: (1) 14 curated tasks that provide standardized evaluation across diverse manipulation categories; (2) 8 reconstructed tasks, where real-world benchmarks are faithfully reconstructed to enable sim-to-real correlated evaluation; and (3) generated tasks, which are automatically generated by recombining assets to support scalable and diverse evaluation through layout generation introduced in Sec.~\ref{sec:layout}.
}

\begin{figure}
    \centering
    \includegraphics[width=1\linewidth]{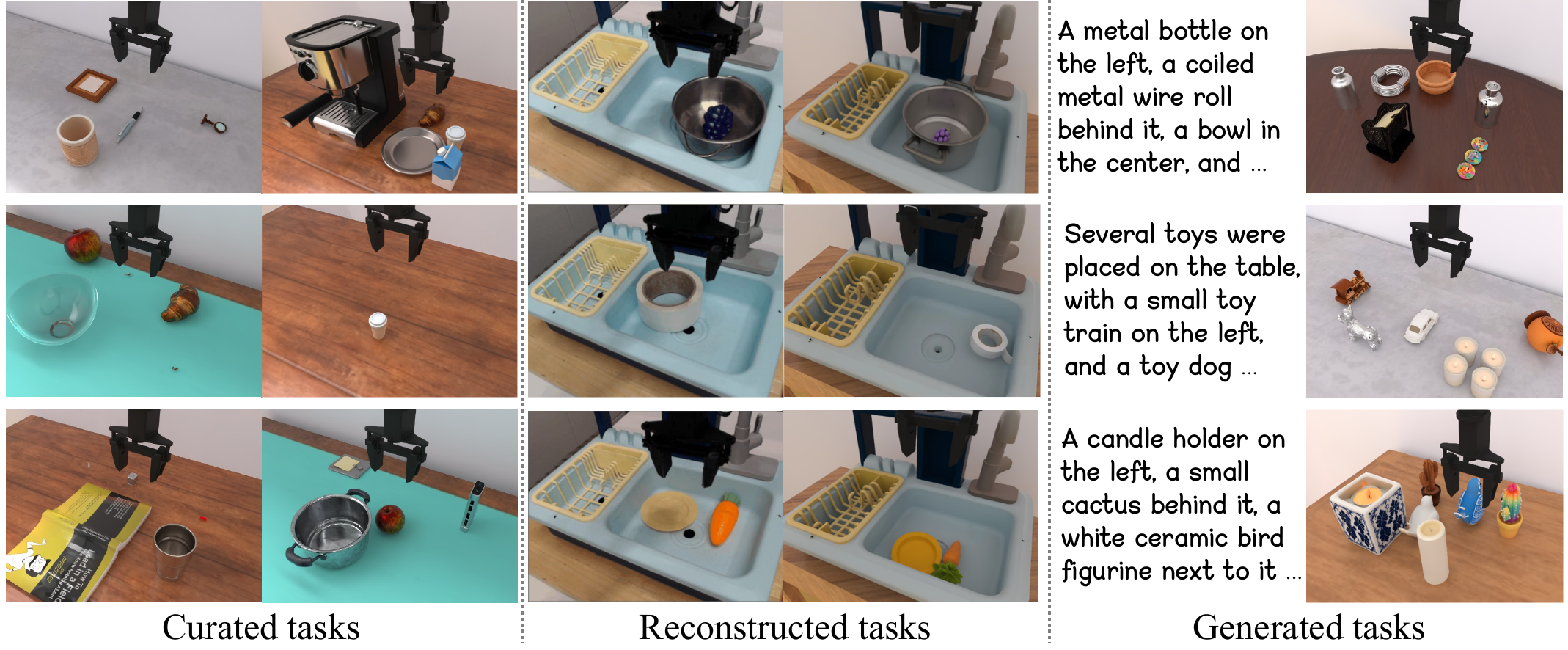}
    \caption{Tasks overview. Our benchmark comprises curated tasks, reconstructed tasks reconstructed from real-world scenes, and generated tasks automatically constructed from descriptions via layout generation.}
    \label{fig:placeholder}
\end{figure}

\subsection{Evaluation Metrics}
For basic tasks, we determine success by monitoring the spatial trajectories and physical states of target objects. For long-horizon tasks, where final-state evaluation is inadequate, we employ Qwen-3-VL~\citep{bai2025qwen3vltechnicalreport} to analyze the execution video. The model is prompted to assess the trajectory based on both \textit{functional success} (the terminal state) and procedural correctness (the logical sequence of actions). This automated evaluation, denoted as Agent Score (AS), provides a robust measure of performance for abstract and multi-step objectives.

\section{Experiment}

In this section, we first demonstrate the quality of our assets in Sec.~\ref{sec:mat_compare}, including comparison of our MLLM-driven material retrieval with the current state-of-the-art material generation methods. Then we evaluate the correlation between simulation and real-world evaluation in Sec.~\ref{sec:correlation}. Last, we evaluate popular VLAs in our simulation in Sec.~\ref{sec:eval_vla}.

\subsection{Asset Quality} \label{sec:mat_compare}

We compare our approach against MaterialMVP \citep{he2025materialmvpilluminationinvariantmaterialgeneration}, Hunyuan3D-v3.1, and Make It Real \citep{fang2024makeitrealunleashinglargemultimodal}. As shown in Fig. \ref{fig:mat_comparison}, our method excels in visual fidelity, capturing fine details like metallic scratches and realistic weathering. Conversely, Hunyuan3D-v3.1 produces simplistic textures with noticeable light baking, while MaterialMVP struggles with high-fidelity generation and lighting-material decoupling. Although Make It Real can enhance albedo, it fails to remove light-baking artifacts and causes structural discontinuities during retrieval due to imprecise segmentation. In contrast, our agent-based inspection ensures robust segmentation. Qualitative comparisons with RoboTwin \citep{chen2025robotwin20scalabledata} and ManiTwin \citep{wang2026manitwinscalingdatagenerationreadydigital} (Table \ref{tab:com_with_datasets_qualitative}), evaluated via VLM-S (Gemini-3.1) and CLIP-S, demonstrate that our dataset achieves superior performance.



\subsection{Sim-Real Correlation} \label{sec:correlation}
Due to the complexities of material properties and lighting conditions, a policy that performs well in simulation may underperform in a real-world environment. To establish a reliable evaluation framework, we assess the correlation between simulated and real-world performance. We constructed simulated counterparts for tasks in BridgeDataV2 \citep{walke2024bridgedatav2datasetrobot} and performed multiple trials for each task. The mapping between simulated and physical environments is detailed in Table \ref{tab:sim2real_corr} (with further specifics in the supplementary material). We report the average success rates for both settings and employ the Pearson Correlation coefficient $r$ to quantify the sim-to-real gap. As shown in Tab.~\ref{tab:sim2real_corr}, the strong correlation confirms the high alignment between our simulation-based evaluation and real-world outcomes. Consequently, we maintain consistent rendering and lighting configurations across all scenes to ensure evaluation reliability. 


\begin{table}[htbp]
    \centering
    \begin{minipage}{0.35\textwidth}
        \centering
        \caption{Qualitative comparison among asset datasets. We randomly sample 100 assets, and leverage Gemini-3.1 to valuate the visual quality score, denoted as VLM-S. We also report the CLIP image-to-text consistency score (CLIP-S).}
        \label{tab:asset_com_score}
        \footnotesize
        \begin{tabular}{lcc}
        \toprule
            & VLM-S$\uparrow$ & CLIP-S$\uparrow$\\
        \midrule
            Ours & \sota{55.35} & \sota{25.20}\\
            RoboTwin & 45.66 & 21.35\\
            ManiTwin & 38.27 & 20.75 \\
        \bottomrule
        \end{tabular}
        \label{tab:com_with_datasets_qualitative}
    \end{minipage}
    \hfill 
    \begin{minipage}{0.63\textwidth}
        \centering
        \begin{threeparttable}
            \caption{Assessment of sim-to-real correlation.}
            \label{tab:sim2real_corr}
            \footnotesize
            \renewcommand\arraystretch{1.1} 
            \begin{tabular}{llccc}
            \toprule
            Policy & Task & Ours & Simpler & Real world \\ \midrule
            \multirow{4}{*}{Octo} & Put spoon/towel & 0.5 & 0.3 & 0.417\\
             & Put carrot/plate & 0.2 & 0.0 & 0.083 \\
             & Eggplant in pot & 0.2 & 0.0 & 0.1\\
             & Eggplant in basket & 0.5 & 0.7 & 0.433 \\ \cmidrule(lr){2-5}
             & \textbf{Correlation ($r\uparrow$)} & \sota{0.9988} & 0.8860 & -- \\ \midrule
            \multirow{5}{*}{OpenVLA} & Carrot/plate v2 & 0.7 & 0.1 & 0.8\\
             & Eggplant in pot & 0.8 & 0.2 & 1.0\\
             & Lift white Tape & 0.1 & 0.3 & 0.1\\
             & Lift battery & 1.0 & 0.0 & 0.7\\
             & Grapes out of pot & 0.2 & 0.0 & 0.4\\ \cmidrule(lr){2-5}
             & \textbf{Correlation ($r\uparrow$)} & \sota{0.8496} & -0.2712 & -- \\
            \bottomrule
            \end{tabular}
        \end{threeparttable}
    \end{minipage}
\end{table}


\subsection{Evaluation of VLAs} \label{sec:eval_vla}
We evaluate several mainstream VLAs in our simulation, including Octo~\citep{octomodelteam2024octoopensourcegeneralistrobot}, OpenVLA~\citep{kim2024openvlaopensourcevisionlanguageactionmodel}, and X-VLA~\citep{zheng2025xvlasoftpromptedtransformerscalable}. To further evaluate the generalization ability on out-of-distribution (OOD) tasks, we sample six tasks. We test all the policies on the six tasks, each categorized into varying difficulty levels. Specifically, while keeping the source and target objects fixed, we systematically introduce challenges such as randomized object positioning, the inclusion of distractor background objects, and the integration of multi-step instructions. Detailed experimental setups are provided in the supplementary material. These scenarios encompass a diverse array of actions, objects, and material properties. Although these VLAs perform well on tasks similar to their training distributions (Sec.~\ref{sec:correlation}) and exhibit a basic ability to process unseen scenes and instructions, their performance remains suboptimal on most OOD tasks. Notably, the success rates drop drastically when background distractors are introduced or when instructions become increasingly complex.

\begin{table}[htbp]
    \vspace{-3mm}
    \centering
    \footnotesize
    \caption{Evaluation of the mainstream policies in our simulation. We report the average success rate.. ``lv.x'' means the different difficulty level, and the detailed description is available in the supplementary. 
    ``LH'' means a long-horizon task. AS denotes Agent Score. We repeat every task for each policy five times and report the average success rate. We define a ``half-success'' as a trial where the object is successfully grasped but fails to reach the target position.}
    \begin{tabular}{ccccccccccccc}
    \toprule
                & \multicolumn{3}{c|}{Pick up paper cup} & \multicolumn{2}{c|}{Put apple in pot} & \multicolumn{2}{c|}{Put bread in bowl} & \multicolumn{2}{c|}{Put cup near book} & \multicolumn{2}{c|}{Open drawer} & \multicolumn{1}{c}{LH} \\
    \midrule
          & lv.1 & lv.2 & lv.3 & lv.1 & lv.2 & lv.1 & lv.2 & lv.1 & lv.2 & lv.1 & lv.2 & AS$\uparrow$\\
    \midrule
         Octo-base & 0.2 & 0.0 & 0.1 & 0.0 & 0.0 & 0.0 & 0.0 & 0.0 & 0.0 & 0.0 & 0.0 & 2.0 \\
         Octo-small & 0.0 & 0.0 & 0.0 & 0.2 & 0.2 & 0.1 & 0.0 & 0.2 & 0.0 & 0.0 & 0.0 & 3.0 \\
         OpenVLA & 0.4 & 0.4 & 0.0 & 0.4 & 0.0 & 0.2 & 0.0 & 0.3 & 0.0 & 1.0 & 0.4 & 5.5 \\
         X-VLA & 0.3 & 0.0 & 0.0 & 0.2 & 0.1 & 0.0 & 0.0 & 0.0 & 0.0 & 0.0 & 0.0 & 4.0 \\
    \bottomrule
    \end{tabular}
    \label{tab:placeholder}
\end{table}

\vspace{-3mm}

\subsection{Discussion and Limitations}
Although we design different task types, including long-horizon tasks, the diversity of tasks is still insufficient compared to the real-world demand. We leave this as future work because we mainly focus on the high-quality assets and the sim-to-real correlation in this paper. We provide two embodiments, Google Robot and WidowX, which are able to achieve most single-arm tasks. Increasing the number of embodiments and evaluating more policies in our simulation is also an important direction for the future.

\section{Conclusion}

In this paper, we introduce VISER, a visually realistic benchmark designed for the reliable evaluation of robot manipulation in simulation. By analyzing the visual domain gap between simulation and the real world, we identify shadows and specular as critical factors impacting the performance of VLA models, thereby underscoring the importance of high-quality PBR materials. To bridge this gap, we develop an MLLM-driven material retrieval and refinement approach to construct a high-quality asset dataset. Powered by these assets and photorealistic rendering, VISER enables scalable and reliable simulation with a demonstrated strong sim-to-real correlation. Through evaluations of VLAs, our benchmark yields valuable insights into their current capabilities and bottlenecks. Future work includes enhance the diversity of tasks and include more embodiments.

\bibliographystyle{plainnat}
\bibliography{reference}


\appendix
\clearpage
\section{Technical appendices and supplementary material}

\subsection{3D Asset Dataset Overview}
\begin{figure}[htbp]
    \centering
    \begin{minipage}[t]{0.48\linewidth}
        \centering
        \includegraphics[width=\linewidth]{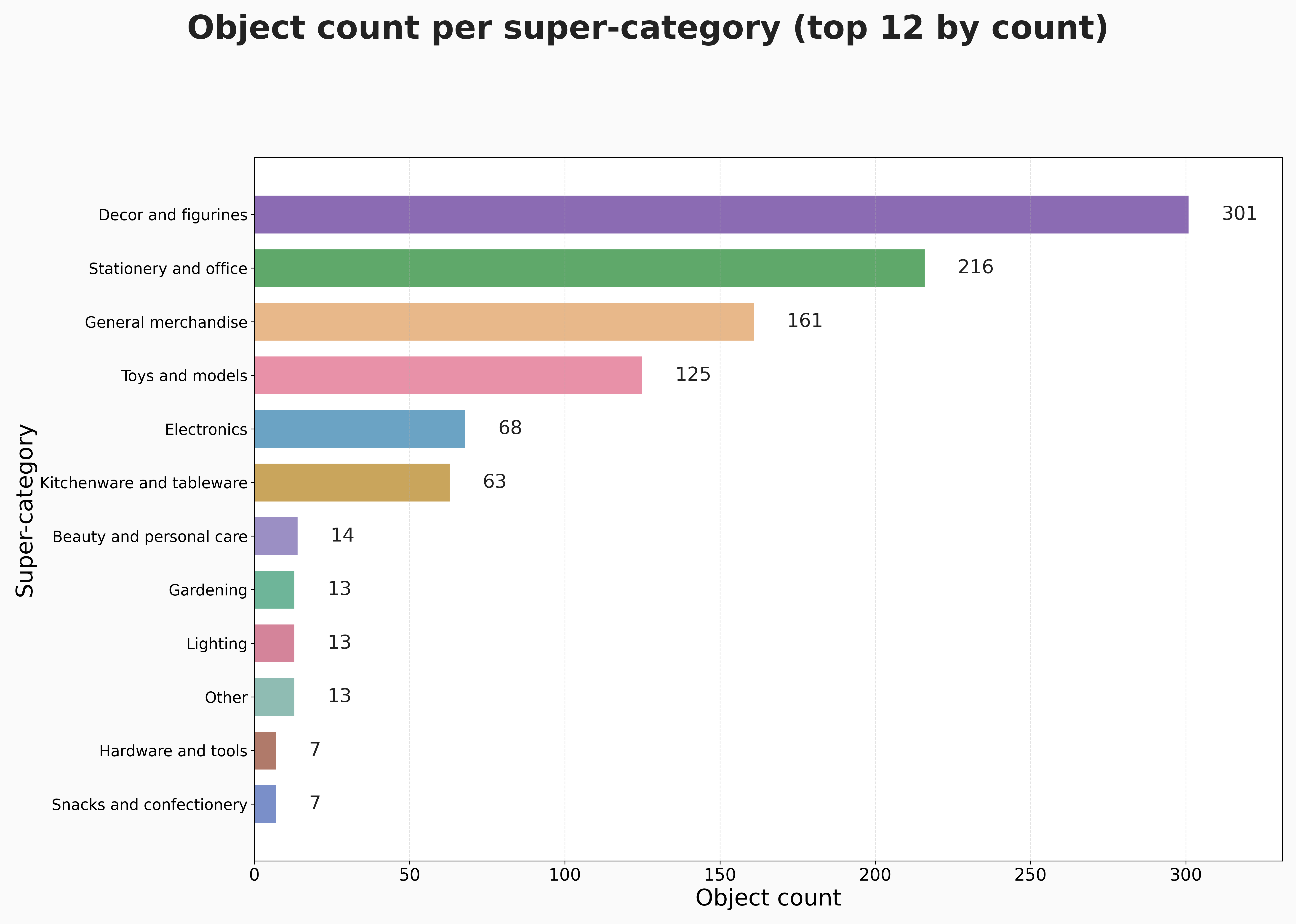}
        \caption{Object count per super-category.}
        \label{fig:supp_count}
    \end{minipage}
    \hfill 
    \begin{minipage}[t]{0.48\linewidth}
        \centering
        \includegraphics[width=\linewidth]{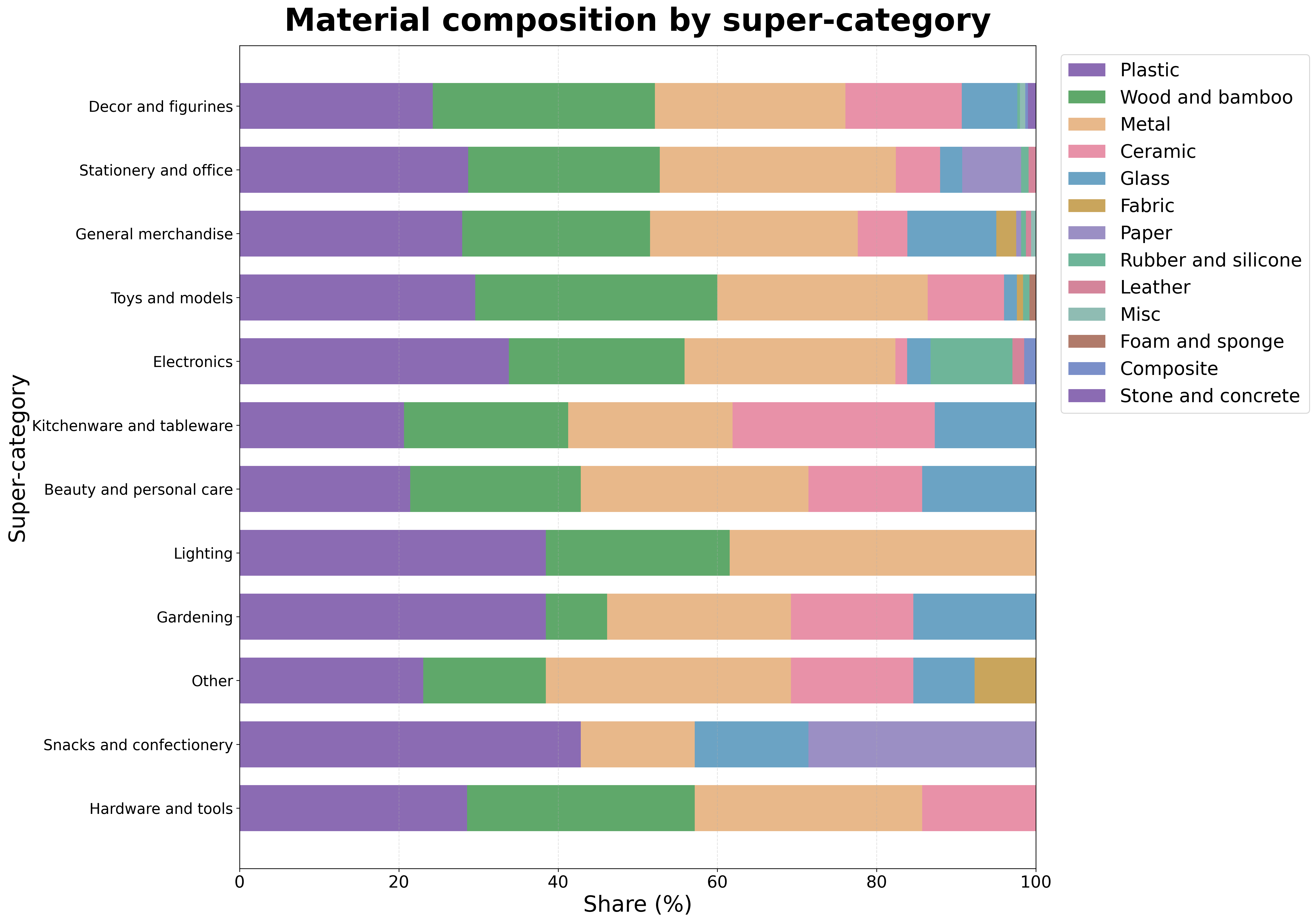}
        \caption{Material composition per super-category.}
        \label{fig:supp_material_composition}
    \end{minipage}
\end{figure}

We present the overview of our 3D asset dataset in Fig.~\ref{fig:supp_count} and Fig.~\ref{fig:supp_material_composition}. Our dataset covers common daily objects and ensures diverse material types.



\subsection{Methodology Detail}

\paragraph{Preliminary asset generation.}
We first leverage GPT-4o to generate a list of common tabletop objects, including categories, names, and prompts. Then we use Tencent HUnyuan3D-v3.0 to generate preliminary geometry and texture. To ensure high-quality geometry, we set triangle numbers to 500K. We use GPT-4o-mini to estimate every object's scale, and calculate the scaling factor according to the estimation and the bounding box. Following similar approaches, we estimate the density for each object. We use CoACD~\citep{Wei_2022} to generate collision geometry. For transparent objects, which generative approaches are able to generate, we set the index of refraction (IOR) and transmission manually.

\begin{figure}[htbp]
    \centering
    \includegraphics[width=1\linewidth]{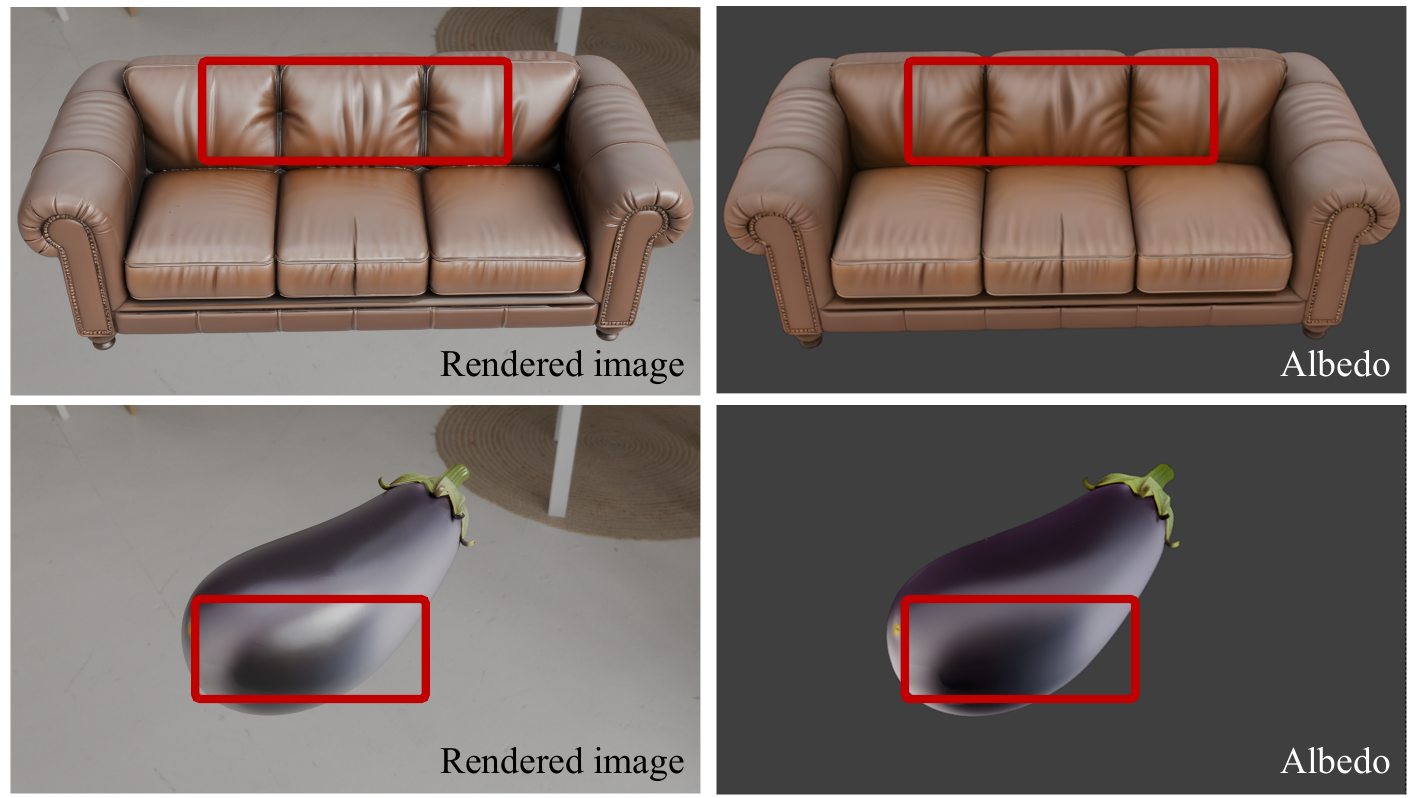}
    \caption{3D generation approaches tend to bake light into the materials. Clean PBR materials ought to disentangle with light.}
    \label{fig:light_baking}
\end{figure}

\paragraph{Material Quality Inspection.}
We observe that most 3D generation approaches struggle to generate clean PBR materials without light baking, as shown in Fig.~\ref{fig:light_baking}. We generate the objects using Hunyuan3D-v3.1~\citep{lai2025hunyuan3d25highfidelity3d}. When the light is mistakenly baked into materials, the object exhibits physically incorrect specular highlights when re-lit under different lighting conditions. To address this issue, we leverage Gemini-3 to inspect each object along with its material. We render an image under a fixed environment light map, and then input it along with the albedo map to Gemini. We observe that a strong VLM can distinguish whether the material is baked light. 

\subsection{Experiment setups}

\begin{figure}[htbp]
    \centering
    \includegraphics[width=1\linewidth]{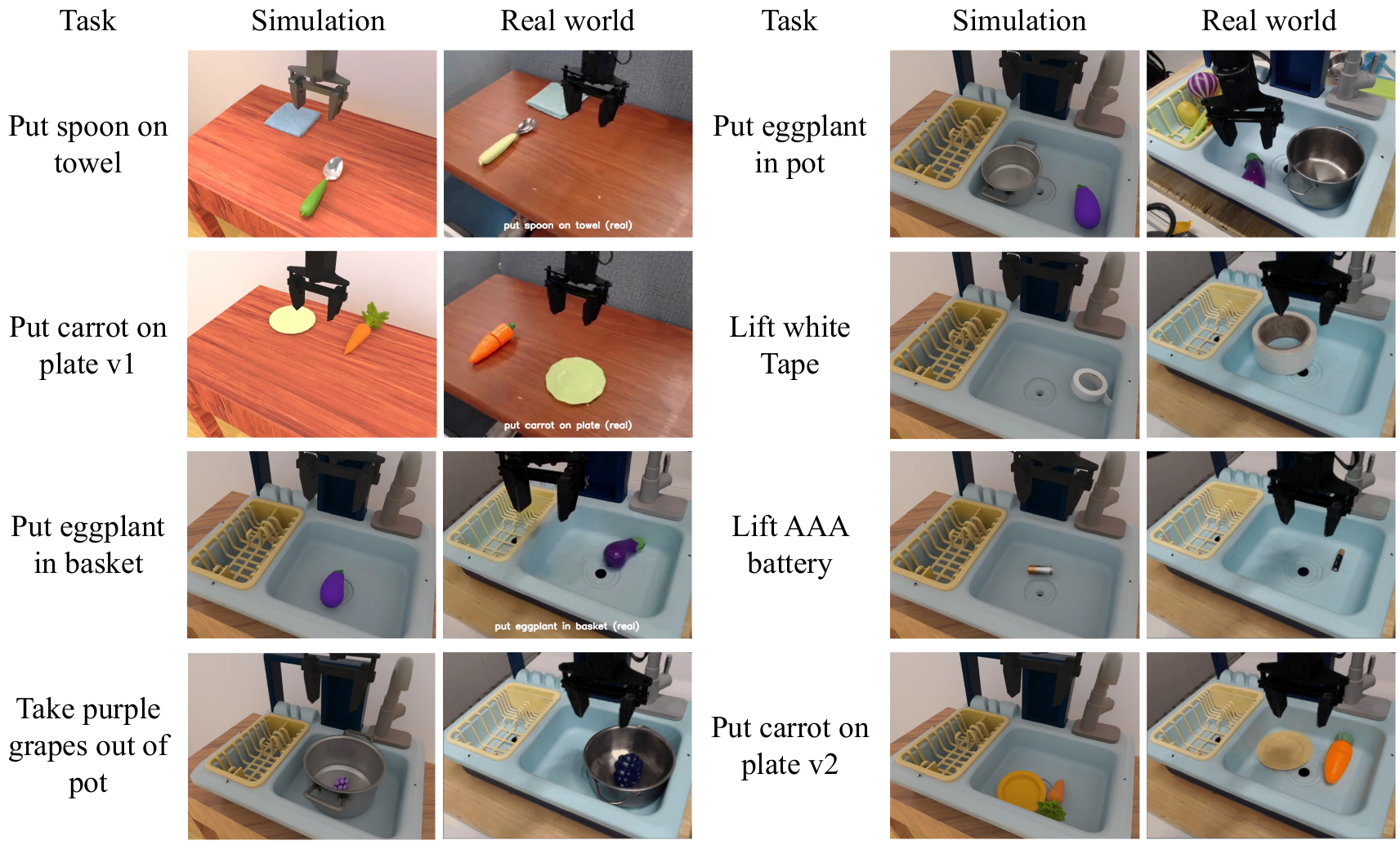}
    \caption{The simulation environment and corresponding real-world scene. }
    \label{fig:sim_real_sup}
\end{figure}

\paragraph{Assessment of sim-real correlation.} We reconstruct the scenes in our simulation according to the real-world BridgeDatav2 scenes. We use Tencent Hunyuan3D-v3.1 to reconstruct the objects through image-3D or text-3D and place them in the scene randomly in a fixed region. The simulation reconstructed scenes and corresponding real-world scenes are shown in Fig.~\ref{fig:sim_real_sup}.

\begin{figure}[htbp]
    \centering
    \includegraphics[width=0.5\linewidth]{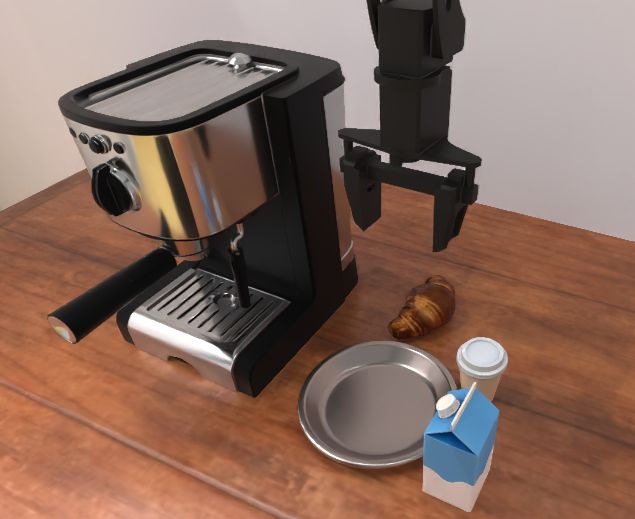}
    \caption{Long-horizon task. Instruction is ``prepare breakfast''.}
    \label{fig:long-horizon}
\end{figure}

\paragraph{Difficulty levels of VLA evaluation.} To evaluate the generalization ability of VLAs, we categorize tasks into different difficulty levels, and the detailed setups are shown in Tab.~\ref{tab:level}. For long-horizon tasks, we use an abstract description rather than explicit instructions to evaluate VLAs' capacity to understand high-level intent and perform multi-step causal reasoning. Unlike primitive tasks that rely on direct spatial mapping, these goals require the model to autonomously decompose a complex objective into a coherent sequence of sub-tasks, thereby testing its ability to generalize its grounding and planning capabilities in dynamic, open-ended environments. As illustrated in Fig.~\ref{fig:long-horizon}, the ``prepare breakfast'' task necessitates that the VLA possesses the contextual knowledge to identify the constituent items, e.g., bread, milk, and coffee. Furthermore, it requires the model to decompose this high-level goal into a sequence of actionable sub-tasks, such as placing the bread on the plate and positioning the cup beneath the coffee machine.

\begin{table}
    \centering
    \small
    \caption{Setups and instructions of different levels.}
    \begin{tabular}{cccc}
    \toprule
        Task & lv.1 & lv.2 & lv.3 \\
    \midrule
        Pick paper cup & only paper cup & w/ kettle and teabox & place the cup behind the kettle\\
        Put apple in pot & only apple and pot & w/ extra 3 bg objects & -- \\
        Put bread in bowl & only bread and bowl & w/ extra 3 bg objects & -- \\
        Put cup near book & only cup and book & w/ extra 3 bg objects & -- \\
        Open drawer & Open middle drawer & Open middle and top drawer & -- \\
        LH & prepare breakfast & -- & -- \\
    \bottomrule
    \end{tabular}
    \label{tab:level}
\end{table}

\subsection{Prompt List}
We show the prompts used in our pipeline, including material-aware part segmentation, material retrieval, and annotation generation.

\begin{tcolorbox}[
  breakable,
  colback=gray!5,
  colframe=black!20,
  title={Prompt-1: Material-aware part segmentation.},
  label={box:prompt_1}
]
\ttfamily\small
Role:
You are an expert in industrial design and material analysis. Multiple images show the SAME 3D object from different camera viewpoints (clay / albedo-style renders).

STEP 1 — DECIDE THE GLOBAL PART LIST FIRST (do this before writing any view output)

Look at ALL the images together and decide ONE fixed set of parts for the whole object.
Rules for choosing the global part list (MATERIAL-FIRST, not function-first):
- **Primary rule: one part = one bulk / surface material class** you would use for PBR texturing
  (e.g. one brushed-metal shell vs. one matte-plastic grip). Split **only** when two regions
  clearly use **different** materials (metal vs. plastic vs. rubber vs. glass, etc.).
- **Do NOT split by functional piece if the material is the same:** e.g. on a kettle or teapot,
  **spout, neck, and main vessel** that are one continuous metal stamping → **a single part**
  with one short name (e.g. "body" or "vessel" — not separate "spout" and "body"). Same for a
  pot: rim + wall + base sharing one metal → one part for that shell.
- **Hard rule — kettles, teapots, pots, pitchers, jars, vases, jugs, coffee pots:** If the
  **spout, rim/lip, side wall, and bottom/base** are the **same** metal or ceramic as the main
  liquid container (one stamped or spun shell), they MUST be **one** part — not separate names
  like "spout", "base", "neck" alongside "body". Use a **single** name (e.g. "body") and bbox(es)
  that cover the whole visible shell (multiple bbox entries for one part are OK).
  **Do not** use geometry-only labels (spout, base, rim) as separate parts when `material` would
  be the same as the main vessel.
- **Anti-pattern (wrong):** `["body", "spout", "handle", "lid", "base"]` when spout, main wall,
  are one continuous metal — merge spout+wall into one part (e.g. "body").
  **Typical** kettle/teapot target: **2–4** parts (metal shell as one part; handle if different
  material; lid if a separate piece you must segment; small knob only if different material).
- **Ignore geometric sub-features** within one material: ears, tail, limbs of uniformly furred
  plastic/metal — still one part per material, not one part per limb.
- If the **same** material appears as **several separate islands** (e.g. two side handles),
  keep **one** part name for that material and list **one bbox rectangle per visible island**
  in the `bbox` array (multiple [[ymin, xmin, ymax, xmax], ...] entries are allowed).
- Aim for **2–5** parts for most objects. Avoid "everything = 1 part" and avoid splitting only
  because the object has many named features (spout, lid knob, …) when they share one material.
- Short, stable English **name** per part (e.g. "body", "handles", "lid"). **Do not** require the
  name to mention material — put material class in the `material` field only.

STEP 2 — APPLY THE SAME PART LIST TO EVERY VIEW

For every view output EXACTLY the same part names from Step 1, in the same order.
Granularity MUST be identical across views — never use a coarser or finer breakdown for one view.
The only allowed exception: omit a part for a specific view if that part is completely occluded
in that view (do not invent a bbox for an invisible surface).

Additional constraints:
- Classify **material type** using object category, shape, and typical manufacturing (metal vs
  plastic vs rubber vs ceramic). You may use diffuse appearance as **weak** evidence (dull vs shiny regions) but must not rely on exact RGB; stay consistent with real-world product builds.
- DO NOT split a part just because it is partially occluded — keep the same name and draw the
  bbox around the visible portion only.
- bbox covers the visible pixels of the part in THAT view; it is fine for bboxes of different
  parts to overlap slightly.

Per part fields:
- name: the stable identifier chosen in Step 1 (reuse verbatim across all views).
- material: e.g. Ceramic, Metal, Plastic, Wood, Fabric, Fur, ...
- description: short phrase describing the appearance of this part in THAT view.
- bbox: [[ymin, xmin, ymax, xmax], ...] with integers in 0..1000 (1000×1000 grid convention).

Output STRICTLY valid JSON only (no markdown fences), schema:
{
  "views": [
    {
      "image\_stem": "<filename stem exactly as given, without extension>",
      "parts": [
        {
          "name": "...",
          "material": "...",
          "description": "...",
          "bbox": [[ymin, xmin, ymax, xmax]]
        }
      ]
    }
  ]
}

The "image\_stem" MUST match one of the stems announced before each image.
\end{tcolorbox}

\begin{tcolorbox}
[
  breakable,
  colback=gray!5,
  colframe=black!20,
  title={Prompt-2: Material retrieval from PBR library.},
  label={box:prompt_2}
]
\ttfamily\small
You assign PBR materials from a fixed library folder.

Object part:
- part\_name: {part.get("name", "")}
- dmx\_material (coarse): {part.get("material", "")}
- appearance notes: {part.get("description", "")}
{oc\_section}
Library category (folder): {category}

Pick exactly ONE material id from the list below. Prefer visual plausibility (color finish,
roughness cues from the name/description) for this object part. The material must fit the
object’s real-world role implied by the part notes and whole-object context (e.g. consumer
cookware vs random industrial sheet metal). Do not invent ids.

Candidates:
{catalog}

Respond with STRICTLY valid JSON only, no markdown:
{{"chosen\_material\_id": "<one id from the list exactly>"}}
\end{tcolorbox}

\begin{tcolorbox}
[
  breakable,
  colback=gray!5,
  colframe=black!20,
  title={Prompt-3: Density estimation.},
  label={box:prompt_3}
]
\ttfamily\small
You are estimating bulk density for physics simulation.

Input: a shaded 3D render that may show PBR materials/textures. Use geometry, proportions, apparent materials, and real-world function.

Task: Give one representative bulk density in kg/m³ for the object as it would exist in the real world (include hollow thin-walled items).

Calibration examples:
- Water or many solid plastics / woods (order of solid household parts): around 1000 kg/m³ (use ~950–1200 when appropriate).
- Empty thin-walled aluminum beverage can (mostly air inside): roughly tens to low hundreds kg/m³ (e.g. ~50–150).
- Solid metals: much higher (e.g. aluminum ~2700, steel ~7800).

Respond with ONLY valid JSON, no markdown:
{"density": <number>, "notes": "<one short sentence>"}
\end{tcolorbox}



\end{document}